\documentclass{article}

\usepackage{arxiv}

\usepackage[utf8]{inputenc} 
\usepackage[T1]{fontenc}    
\usepackage{hyperref}       
\usepackage{url}            
\usepackage{booktabs}       
\usepackage{amsfonts}       
\usepackage{nicefrac}       
\usepackage{microtype}      
\usepackage{lipsum}		
\usepackage{graphicx}
\usepackage{natbib}
\usepackage{doi}

\usepackage{amsmath}
\usepackage[T1]{fontenc}
\usepackage[dvipsnames,table,xcdraw]{xcolor}
\usepackage{array}  
\usepackage{graphicx}
\usepackage[most]{tcolorbox}

\newtcolorbox{outerBox}[1]{
    breakable, enhanced,
    colback=black!5!white,
    colframe=black!75!white,
    title={Box {\bf #1}},
    top=5pt
}

\newtcolorbox{twocolBox}{
    sidebyside, sidebyside align=top,
    enhanced,
    arc=0pt,
    colback=black!5!white,
    colframe=black!75!white,
    boxsep=1mm,
    notitle,
    oversize,
    nobeforeafter,
    frame hidden,
}

\usepackage{array,tabularx,lipsum}
\newtcolorbox{MyBox}[1][]{%
    enhanced,
    boxsep=5pt,
    fonttitle=\bfseries\large,
    fontupper=\normalsize\ttfamily,
    colback=yellow!10!white,
    colframe=red!50!black,
    colbacktitle=red!30!white,
    coltitle=black,
    before title={\noindent},
    #1
}%

\title{More Than the Final Answer: Improving Visual Extraction and Logical Consistency in Vision–Language Models

}

\author{ 
Hoang Anh Just\textsuperscript{1}\thanks{Work done while interning at Adobe Research.},
Yifei Fan\textsuperscript{2}, Handong Zhao\textsuperscript{2}, Jiuxiang Gu\textsuperscript{2}, Ruiyi Zhang\textsuperscript{3},\\ \textbf{Simon Jenni\textsuperscript{2}, Kushal Kafle\textsuperscript{2}, Ruoxi Jia\textsuperscript{1}\thanks{Equal advising.}, Jing Shi\textsuperscript{2}\footnotemark[2]}\\
	\textsuperscript{1}Virginia Tech, \textsuperscript{2}Adobe Research, \textsuperscript{3}Apple\\
}

\renewcommand{\shorttitle}{Improving Visual Extraction and Logical Consistency in Vision–Language Models}

\hypersetup{
pdftitle={More Than the Final Answer: Improving Visual Extraction and Logical Consistency in Vision–Language Models},
pdfsubject={cs.LG, cs.CV, cs.CL},
pdfauthor={Hoang Anh Just, Yifei Fan, Handong Zhao, Jiuxiang Gu, Ruiyi Zhang, Simon Jenni, Kushal Kafle, Ruoxi Jia, Jing Shi },
pdfkeywords={VLM, reward, vision language model, description, RLVR, verifiable rewards, perl, visual extraction},
}

\begin{document}
\maketitle

\begin{abstract}
	Reinforcement learning from verifiable rewards (RLVR) has recently been extended from text-only LLMs to vision–language models (VLMs) to elicit long-chain multimodal reasoning. However, RLVR-trained VLMs still exhibit two persistent failure modes: inaccurate visual extraction (missing or hallucinating details) and logically inconsistent chains-of-thought, largely because verifiable signals supervise only the final answer. We propose \textbf{PeRL-VL} (\textbf{Pe}rception \& \textbf{R}easoning \textbf{L}earning for \textbf{V}ision–\textbf{L}anguage Models), a decoupled framework that separately improves visual perception and textual reasoning on top of RLVR. For perception, PeRL-VL introduces a VLM-based description reward that scores the model’s self-generated image descriptions for faithfulness and sufficiency. For reasoning, PeRL-VL adds a text-only Reasoning SFT stage on logic-rich chain-of-thought data, enhancing coherence and logical consistency independently of vision. Across diverse multimodal benchmarks, PeRL-VL improves average Pass@1 accuracy from 63.3\% (base Qwen2.5-VL-7B) to 68.8\%, outperforming standard RLVR, text-only reasoning SFT, and naive multimodal distillation from GPT-4o.
\end{abstract}


\section{Introduction}

Reinforcement learning from verifiable rewards (RLVR) has recently emerged as a powerful recipe for “making models think,” exemplified by DeepSeek-R1~\cite{guo2025deepseek}, which uses rule-based verifiable signals (e.g., unit tests, exact-answer checks) to elicit long-chain reasoning.
This paradigm is now being actively extended from text-only LLMs to vision–language models (VLMs)~\cite{shen2025vlmr1,huang2025visionr1} for multimodal tasks, showing that verifiable outcome rewards can significantly boost visual reasoning and chain-of-thought (CoT) performance. 
However, despite these advances, we observe that RLVR-trained VLMs still exhibit two stubborn failure modes in practice: (1) visual extraction errors, where the model hallucinates objects, misses crucial visual attributes, and (2) logical reasoning errors, where the final answer may be correct but the underlying CoT is inconsistent or self-contradictory even when the perception is accurate. We hypothesize that both issues stem from the outcome-only nature of standard RLVR: the verifier typically inspects only the final answer, which encourages policies that tolerate spurious or shortcut reasoning, leading to hallucinated processes and "reward hacking".

\begin{figure}
    \centering
    \includegraphics[width=0.59\linewidth]{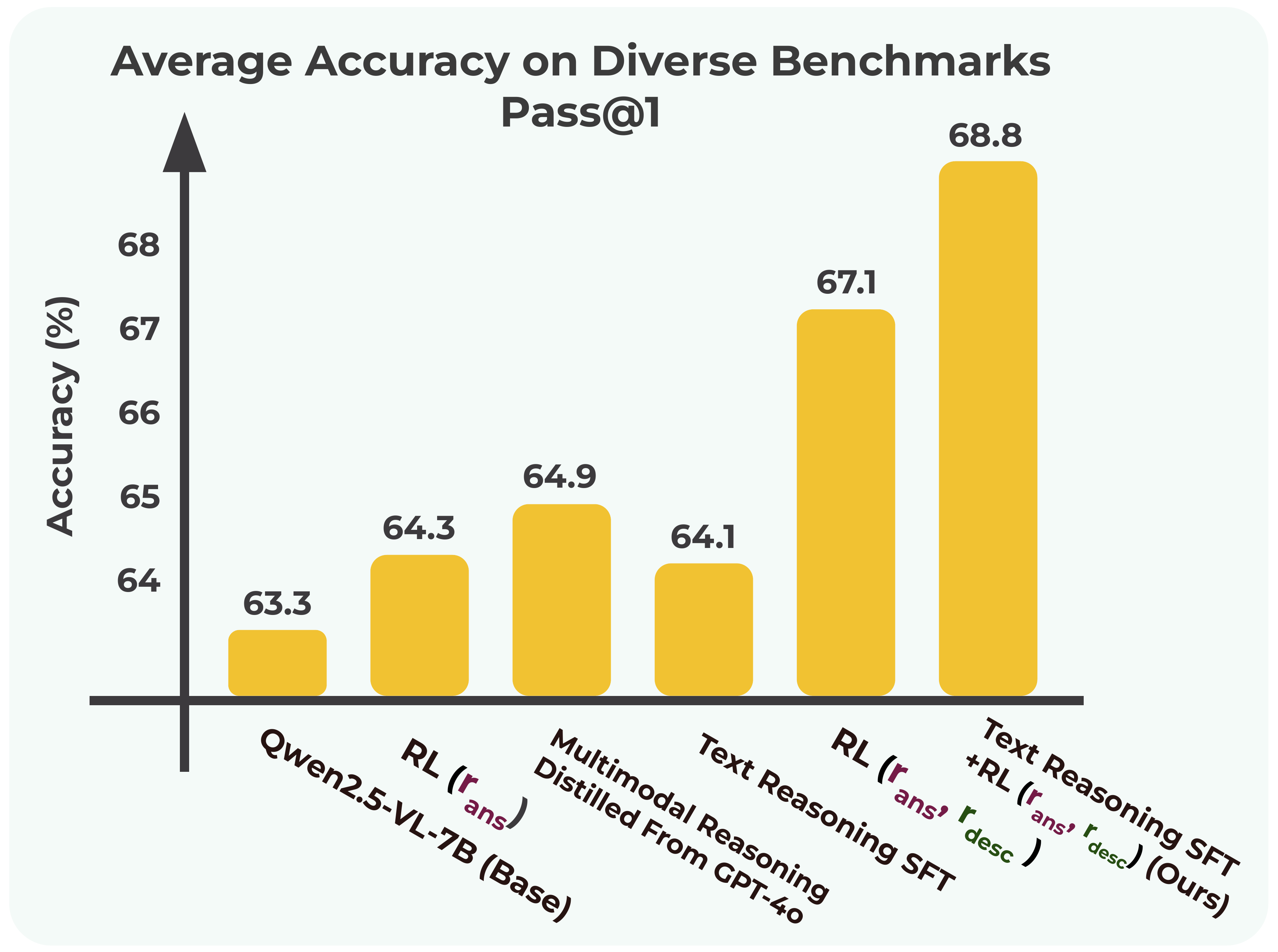}
    \caption{Average performance comparison between different methods across benchmarks. RL($r_{\text{ans}}$) denotes verifiable RL with answer and format rewards. RL($r_{\text{ans}}$, $r_{\text{desc}}$) further adds description reward in addition to verifiable rewards. Reporting Pass@1 accuracy.} 
    \label{fig:teaser}
\end{figure}

Motivated by these observations, we explicitly decouple visual extraction and textual reasoning instead of treating VLM ``thinking” as a single block. 
We view VLM inference as two stages: a perception stage, where the model generates a sufficiently detailed, faithful description of the image, and a reasoning stage, where a text-only chain-of-thought operates on that description to produce the answer. 
A naive approach is to directly distill descriptions and reasoning from a strong multimodal model (e.g., GPT-4o) via Supervised fine-tuning (SFT). However, we find this yields only a small performance gain, as shown in Fig.~\ref{fig:teaser}. Hence, we propose PeRL-VL, which comprises two dedicated, decoupled modules --  one each for perception and reasoning.

First, PeRL-VL enriches the RLVR signal with explicit supervision for the perception stage. The policy model must produce its own description of the image, and a VLM reward model judges whether this description is both faithful to the visual content and sufficient for solving the task. Using a VLM as a reward also enables natural co-improvement as the reward model itself scales~\cite{liu2025inference}. Moreover, we systematically compare \emph{Aggregated} reward composition (a simple mixture of description and outcome rewards) with a stricter \emph{Conditional} (gated) scheme that only grants a high outcome reward when both perception and answer are correct. The gated design more strongly discourages reward hacking and pushes the model to fix its visual extraction before relying on complex reasoning, leading to substantially better visual extraction quality in PeRL-VL.

Second, PeRL-VL enhances logical consistency by introducing a \emph{Text Reasoning SFT} stage, where the model is trained on text-only logic-rich reasoning data (e.g., OpenThought) via SFT, explicitly guiding it to generate a more coherent, logically consistent CoT. Because this operates purely in text space, it improves the reasoning module's robustness independent of visual perception, cleanly complementing the perception-focused reward in PeRL-VL.

Our contributions are as follows:

\begin{itemize}
    \item We demonstrate the unreliability of purely verifiable rewards and introduce a VLM-judged description reward to improve visual perception.
    
    \item We propose PeRL-VL, a decoupled framework that targets perception with a novel RL reward design and reasoning with a dedicated SFT stage.

    \item We provide a systematic study of reward composition (\emph{Aggregated} vs. \emph{Conditional}), showing that the design of the reward function is a critical factor in mitigating reward hacking.

    \item We present a clear analysis linking improved performance to a reduction in "false positive" rollouts, explaining why different reward designs lead to a better generalization.

    \item We show that our modular framework is highly effective, improving both logical reasoning and visual faithfulness, synergistically.

\end{itemize}

\section{Related Work}
\label{sec:related}

\subsection{SFT-based Post-Training}
SFT has long served as the foundation for adapting pretrained large language models with visual encoders to multimodal reasoning tasks. Early efforts leveraged large-scale multimodal instruction datasets, while recent work increasingly emphasizes structured reasoning supervision. For instance, LLaVA-CoT explicitly trains models to decompose reasoning into staged "see–think–answer" outputs and employs re-tracing to verify consistency and improve accuracy during inference~\cite{xu2024llavacot}. However, SFT alone can induce pseudo-reasoning, where models imitate stylistic reasoning patterns without genuine understanding. As shown by Chen et al., overfitting to expert chain-of-thought traces can hinder downstream reinforcement learning (RL) optimization and lead to brittle reasoning~\cite{chen2025sft_or_rl}.

To alleviate such limitations, newer frameworks integrate preference optimization within the SFT pipeline. InternVL3 incorporates Mixed Preference Optimization (MPO) to align reasoning with both human and verifiable objectives~\cite{zhu2025internvl3}, while InternVL3.5 further introduces Cascade RL for stable convergence and efficient long-context reasoning~\cite{wang2025internvl35}. Similarly, Skywork-R1V2 combines MPO with Group Relative Policy Optimization (GRPO) and selective replay buffers to enhance stability~\cite{wang2025skywork_r1v2}. Parallel work such as VIRAL regularizes internal representations, ensuring that visual embeddings remain consistent with frozen vision encoders to preserve fine-grained perceptual fidelity~\cite{yoon2025viral}. Practical extensions also emerge in production-grade systems—e.g., GLM-4.5V introduces a dynamic “thinking mode” to balance accuracy and latency~\cite{hong2025glm45v}, and Kimi-VL adopts a Mixture-of-Experts design for efficient inference~\cite{kimi2025kimivl}.
Overall, SFT remains an effective initialization step for multimodal reasoning. Yet, it requires carefully designed objectives and curricula to prevent imitation bias, retain perceptual grounding, and prepare models for subsequent RL optimization.

\begin{figure*}[h!t]
    \centering
    \includegraphics[width=0.99\linewidth]{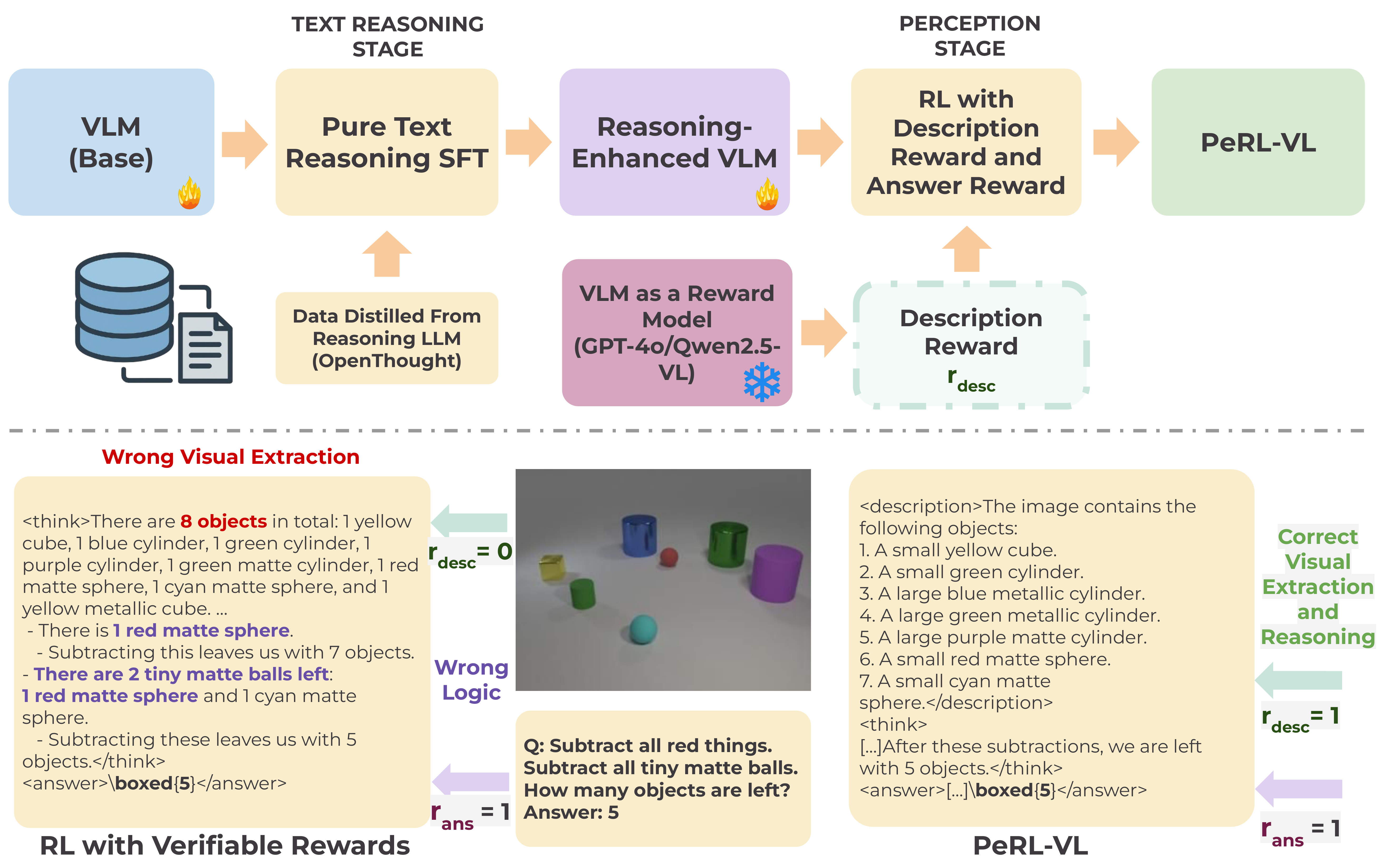}
    \caption{(Above) An overview showing the modular framework of PeRL-VL. The base VLM starts with the Text Reasoning Stage to improve logical reasoning, where it is fine-tuned on reasoning data distilled from a language reasoning model. Then, the model proceeds to the Perception Stage, where it learns to extract visual information faithfully while being supervised by a description reward via reinforcement learning. (Below) An example comparing responses from Verifiable RL and PeRL-VL. The left response shows an error due to incorrect visual extraction and wrong reasoning; however, the model is still rewarded due to a matching final answer. The right response demonstrates improved visual extraction capability and correct reasoning.}
    
    \label{fig:pipeline}
\end{figure*}

\subsection{Reinforcement Learning for VLMs}
Reinforcement learning (RL) has recently become central to enhancing both reasoning depth and visual alignment in VLMs. Unlike SFT—which supervises outputs directly—RL allows models to learn from verifiable signals such as answer correctness, intersection-over-union (IoU), or counting accuracy.
R1-VL introduces step-level accuracy and validity rewards through StepGRPO, ensuring that intermediate reasoning steps are interpretable and consistent~\cite{zhang2025r1vl}. Visual-RFT applies RL-based fine-tuning to grounding and detection tasks with IoU-driven rewards, outperforming SFT in visual generalization~\cite{liu2025visualrft}. VisionReasoner extends this to structured multi-object reasoning, incorporating multi-stage verifiable rewards to achieve strong perception and reasoning performance from limited data~\cite{liu2025visionreasoner}. Similarly, VLM-R1~\cite{shen2025vlmr1} and Vision-R1~\cite{huang2025visionr1} demonstrate that visually-grounded reward functions can yield robust, out-of-distribution generalization.

Beyond static reward designs, curriculum and sample-selection strategies further stabilize RL training. VL-Cogito~\cite{yuan2025vlcogito} employs progressive curriculum RL, dynamically adjusting difficulty and reward weighting. MM-Eureka~\cite{meng2025mmeureka} adopts a two-stage pipeline—using online filtering and curated math data—to enhance multimodal reasoning. MCTS-guided sampling~\cite{wang2025sotawithless} leverages reasoning difficulty to identify high-value rollouts, drastically improving efficiency.
Other frameworks close the SFT–RL loop: OpenVLThinker~\cite{deng2025openvlthinker} alternates between RL refinement and self-generated SFT data, producing iterative reasoning improvement. LMM-R1~\cite{peng2025lmmr1} shows that text-only RL can bootstrap reasoning skills later transferable to multimodal settings, while G1~\cite{chen2025g1} explores embodied RL to jointly evolve perception and reasoning. Large-scale systems such as InternVL3.5~\cite{wang2025internvl35} and GLM-4.5V~\cite{hong2025glm45v} further demonstrate that cascading or hierarchical RL can balance efficiency and long-context reasoning quality.

\subsection{Reasoning and Perception Decomposition Training}
Decomposition of reasoning has recently been proposed as a means to improve supervision, in which the perception stage is often evaluated using proxy signals such as self-reward for caption sufficiency~\cite{li2025self} or CLIP scores~\cite{chen2025perception} for alignment. While valuable, these proxies may not fully capture perception errors. Our work builds on this by proposing a more direct supervision method. We use a VLM as a reward model to explicitly assess the generated description against the image itself for visual extraction errors. This allows for a more direct and high-fidelity evaluation of perception, which we integrate into the perception stage.

\section{Method}

We propose PeRL-VL, a decoupled modular framework for post-training vision-language models that is designed to address their two primary failure modes: flawed logical reasoning and unfaithful visual extraction. First, we enhance the model's logical reasoning capabilities using supervised fine-tuning (SFT). Second, we ground that reasoning in visual evidence by improving perception with reinforcement learning (RL), as illustrated in Figure~\ref{fig:pipeline}.

\subsection{Perception Stage}
\label{sec:perception}
Our goal in the perception stage is to strengthen multimodal reasoning by ensuring that reasoning chains remain supported by observable visual evidence rather than spurious or hallucinated descriptions. 
Our empirical finding that direct SFT distillation of descriptions from a capable VLM yields limited gains motivates our use of RL for this stage (Section~\ref{sec:sft_rl}). We design an RL framework that provides explicit, high-fidelity supervision for the model's perception, built on three key components: a structured output format, VLM-guided rewards, and a systematic study of reward composition.

\subsubsection{Structured Output for Grounded Reasoning}

To facilitate targeted supervision, we require the policy model,$\pi_\theta$, to generate a structured output for any given question-image pair, $Q=(i,q)$. This output follows a "See-Think-Answer" format: $s=\langle c,\, t,\, a \rangle$ with \textless{}description$\text{>}$c$\text{<}$/description$\text{>}$ $\text{<}$think$\text{>}$t$\text{<}$/think$\text{>}$ $\text{<}$answer$\text{>}$a$\text{<}$/answer$\text{>}$, 
where $c$ denotes the model’s visual description comprising faithful and sufficient visual evidence (\emph{perception stage}), $t$ is the chain-of-thought reasoning trace (\emph{reasoning stage}), and $a$ is the final answer (\emph{conclusion stage}). This decomposition makes the model's intermediate steps explicit, enabling simpler external evaluation of its perceptual accuracy.

\subsubsection{RL with VLM-Guided Rewards}
The policy $\pi_\theta$ is optimized using group-relative policy optimization (GRPO)~\cite{shao2024deepseekmath}, a variance-reduced variant of policy-gradient RL. For each question $Q$, we sample $K$ rollouts $\{s_k\}_{k=1}^{K}$, compute scalar rewards $r(Q,s_k)$, and normalize them within the batch to obtain the advantage estimate: $\hat{A}_{\mathrm{grp}}(Q,s_k) = r(Q,s_k) - \frac{1}{K} \sum_{j=1}^{K} r(Q,s_j).
$
The optimization objective is
$
\mathcal{L}(\theta)=\mathbb{E}_{Q}\!\left[\sum_{k=1}^K \hat{A}_{\text{grp}}(Q,s_k)\,\log \pi_\theta(s_k \mid Q)\right] 
- \beta\,\mathrm{KL}\!\left(\pi_\theta(\cdot\mid Q)\,\|\,\pi_{\theta_0}(\cdot\mid Q)\right),
$
where $\pi_{\theta_0}$ is the frozen reference policy.

We define two verifiable rewards:
\begin{itemize}
    \item \textbf{Format reward} $r_{\mathrm{fmt}}(s)=1$ if tags are matched, else $0$.
    \item \textbf{Answer reward} $r_{\mathrm{ans}}(Q,a)=1$ if $a$ matches the ground truth, else $0$.
\end{itemize}
This baseline, a standard Verifible RL, improves structural correctness but remains vulnerable to \emph{pseudo-reasoning}: correct answers achieved via hallucinated or template-based justifications.

To mitigate pseudo-reasoning, we introduce an extra multimodal model $\mathcal{T}$ (e.g., GPT-4o) to provide \textbf{a description reward}, $r_{desc}$, that directly verifies whether the model’s visual description $c$ is both \emph{factually faithful to the image} and \emph{sufficient for solving the task}: 
$
r_{\mathrm{desc}}(Q,c) =1$ if $c$ is a faithful description of the image and provides sufficient information to solve $Q$, else $0$.
This unified evaluation ensures that the model is rewarded only for descriptions that are both visually correct and useful, preventing credit from being assigned for irrelevant but accurate details. We provide prompt templates in Appendix C.

\subsection{Reward Decomposition and Design Variants}
The final reward function combines the verifiable reward signals with the perception reward. We investigate two primary strategies for combining these reward signals.

\paragraph{(A) \emph{Aggregated} rewards.}
All components contribute linearly, providing dense and stable supervision early in training, as the model can receive partial credit for correct answers even if its visual description is flawed: 
$
r(Q,s) = 
\alpha_{\mathrm{fmt}} r_{\mathrm{fmt}} +
\alpha_{\mathrm{desc}} r_{\mathrm{desc}} +
\alpha_{\mathrm{ans}} r_{\mathrm{ans}},
$
where $\alpha$ terms are weighting coefficients with $\sum_i \alpha_i = 1$.

\paragraph{(B) \emph{Conditional} rewards.}
This structure introduces a strict dependency, rewarding a correct answer only if the underlying visual description is verified by the VLM:
$
r(Q,s, \gamma)=
\alpha_{\mathrm{fmt}}r_{\mathrm{fmt}}+ 
\alpha_{\mathrm{ans}}\big[\gamma r_{\mathrm{ans}}+(1-\gamma)(r_{\mathrm{ans}}\cdot r_{\mathrm{desc}})\big].
$
This gating mechanism enforces a causal link between perception and reasoning, discouraging the model from finding "right answers for the wrong reasons." In our experiments, we study three cases, $\gamma=1$ (hard gate), which neglects the description reward and boils down to verifiable rewards RL, $\gamma=0$, which gives a reward if and only if both description and answer are correct, $\gamma=0.5$, a softer gating version, which allows partial rewards in cases where only the answer is correct. We denote the RL setting with reward function $r(Q,s, \gamma)$ as $RL(\gamma)$.

\subsection{Text Reasoning Stage}
\label{sec:logical}
Even with perfect perception, a VLM can falter due to an inconsistent or illogical CoT, as shown in Figure~\ref{fig:pipeline}. To address this, we begin with a text-only logical reasoning SFT stage. Inspired by prior work showing that SFT on text-only reasoning data can improve a smaller model's logical capabilities~\cite{guha2025openthoughts, liu2025acereason, tu2025mlan}, we fine-tune our policy model on a curated, reasoning-rich dataset such as OpenThought~\cite{guha2025openthoughts}, which is distilled from a strong reasoning language model, such as DeepSeek-R1~\cite{guo2025deepseek}.
This stage is performed before the perception-focused RL. By operating purely in the text domain, it equips the model with a stronger, more coherent reasoning backbone, improving its logical reasoning and maintaining logical consistency, independent of any visual input. Afterwards, we proceed to the perception stage to align the textual logical reasoning with correct visual understanding, which leverages improved logical reasoning to improve visual understanding and reasoning.

\begin{table*}[t]
\centering

\resizebox{1.00\linewidth}{!}{
\begin{tabular}{lcccccc}
\toprule
\textbf{Model}                                                                           & \textbf{MathVista} & \textbf{MMMU} & \textbf{MMBench} & \textbf{OCRBench} & \textbf{Hallusion} & \textbf{AVG} \\
\toprule
Base (Qwen2.5-VL-7B)~\cite{bai2025qwen2}                                                            & 57.78           & 48.15          & 82.90                 & 80.20    & 43.80      & 62.57    \\
\midrule
\multicolumn{7}{c}{\itshape Reasoning-Oriented Models }\\
\midrule
ThinkLite-VL~\cite{wang2025sotawithless}                                 & 65.00 & 50.81 & 82.65 & 84.82 & 36.10 & 63.88\\
VL-Cogito~\cite{yuan2025vlcogito}                                    & 61.30 & 51.72 & 83.92 & 84.64 & 46.63 & 65.64\\
R1-ShareVL-7B~\cite{yao2025r1}                                & 66.10 & 52.28 & 83.67 & 85.85 & 44.53 & 66.49\\
PeBR-R1-7B~\cite{chen2025perception}                                   & 70.50 & 51.59 & 83.78 & 86.31 & 49.56 & 68.35\\
Vision-SR1~\cite{li2025self}                                   & 63.50 & 52.80 & 84.30 & 82.37 & 43.20 & 65.23\\
\midrule
\multicolumn{7}{c}{\itshape PeRL: Reasoning Stage Only }\\
\midrule
SFT (GPT-4o~\cite{hurst2024gpt})               &    58.90     & 48.06   &  82.57   &         82.58             & 52.30 & 64.88 \\
SFT (OpenThought~\cite{guha2025openthoughts})                                                          & 67.50 & 48.51 & 81.10 & 82.20 & 41.40 & 64.14\\
\midrule
\multicolumn{7}{c}{\itshape PeRL: Perception Stage Only }\\
\midrule
RL($\gamma=1$) (Verifiable Rewards)                                                & 64.15           & 51.42          & 83.20                 & 85.13    & 37.67    & 64.31      \\
\textbf{RL }(Aggregated Rewards)   (Ours)                                           & 65.45           & 50.80          & 83.20                 & 86.30    & 47.33       & 66.62   \\
\textbf{RL($\gamma=0.5$)}  (Conditional Soft Gate)    (Ours)                                        & 65.80           & 52.09          & 84.15                 & 86.56    & 46.12     & 66.94     \\
\textbf{RL($\gamma=0$)} (Conditional Hard Gate)      (Ours)                                       & \textbf{66.11}  & \textbf{52.11} & \textbf{84.29}        & 86.16    & \textbf{47.22}  & 67.18\\ 
\midrule
\rowcolor{blue!10}
\textbf{PeRL-VL}  (Ours)                            & \textbf{67.05} & \textbf{52.22} & 83.50 & 85.10 & \textbf{55.90} & \textbf{68.75} \\
\bottomrule
\end{tabular}
}
\caption{Performance on benchmarks for models trained with different reward functions or methods, including the base model and SFT trained models. Comparison with open-source baselines. All results are Pass@1.}
\label{tab:mainresult}
\end{table*}

\section{Experiments}

In this section, we empirically validate the PeRL-VL framework. We begin by analyzing the Text Reasoning SFT stage, showing how it improves the model's abstract reasoning capabilities. We then present our main results on the Perception RL stage, demonstrating the superiority of our reward designs over a standard Verifiable RL baseline. This is followed by a deep dive into the training dynamics to explain why our method works. Finally, we showcase the power of the full hybrid PeRL-VL framework, benchmark it against state-of-the-art models, and perform an ablation on VLM reward models.

\subsection{Experimental Setup}

\paragraph{Models and Datasets.}
Our experiments use Qwen2.5-VL-7B-Instruct as the policy model in the main paper. For description supervision, we employ GPT-4o, unless otherwise specified. All models are fine-tuned on a custom 80K-sample dataset aggregated from ThinkLite-VL~\cite{wang2025sotawithless} and VL-Cogito~\cite{yuan2025vlcogito}) to ensure comprehensiveness and quantity (we have observed smaller improvement when training on ThinkLite-VL alone. We refer the reader to Appendix B.) 
\newline
\textbf{SFT.} We perform full-parameter SFT using the LLaMA-Factory framework~\cite{zheng2024llamafactory}. The training is conducted for 3 epochs with a global batch size of 256, a context length of 32k, and a learning rate of $5\times 10^{-6}$.
\newline
\textbf{RL.}
For the RL phase, we use the VERL implementation of GRPO. The policy is trained for 100 steps with a learning rate of $1\times 10^{-6}$, a global batch size of 512, and a context length of 20k. In each step, we generate 16 rollouts per input. To maintain stability, we set a KL-divergence penalty to 0.001. We set $\alpha_{\mathrm{fmt}}$ to 0.1, and $\alpha_{\mathrm{desc}} = \alpha_{\mathrm{ans}} = 0.45$ for the aggregated case.
\newline
\textbf{Structured outputs.}
Our framework requires the student model to generate outputs in a structured ``See-Think-Answer'' format. For the Verifiable RL baseline, we only enforce the ``Think-Answer'' format.
\newline
\textbf{Benchmark Evaluation.}
Model performance is evaluated using the datasets from the VLMEvalKit framework with a sampling temperature of 0.6, a context length of 32k, Qwen2.5-72B-Instruct as a judge, and by reporting \textbf{Pass@1} over $8$ generations. We report performance across a diverse suite of benchmarks to assess different capabilities: OCRBench: OCR-centric visual reasoning, MathVista\_MINI (MathVista): Visual mathematical reasoning, MMMU\_DEV\_VAL (MMMU): Expert-level multidisciplinary QA,  MMBench\_DEV\_EN\_V11 (MMBench): General-purpose multiple-choice evaluation, HallusionBench (Hallusion): Diagnostic benchmark for visual hallucination.
Additional results appear in Appendix B.

\subsection{Comparison to Contemporary Methods }

To contextualize our contributions, we benchmark our model, PeRL-VL, against several strong, publicly available baselines (ThinkLite‑VL~\cite{wang2025sotawithless}, VL‑Cogito~\cite{yuan2025vlcogito}, R1-Share‑VL‑7B~\cite{yao2025r1}, PeBR‑R1‑7B~\cite{chen2025perception}, Vision-SR1~\cite{li2025self}), as shown in Table~\ref{tab:mainresult}. 
Our framework's primary strength is demonstrated on HallusionBench, where our PeRL-VL model achieves a score of $55.90$, a significant improvement over all other compared models. This result underscores our method's superior capability in controlling visual hallucinations. While specialized models such as PeBR-R1-7B attain a higher score on MathVista\_MINI, we could also further improve the reasoning stage by scaling the data at SFT as PeBR-R1-7B does. Furthermore, our models achieve great performance on general benchmarks such as MMMU\_DEV\_VAL and MMBench\_DEV\_EN, reflecting broad improvements to visual grounding and reliability. Crucially, our description reward mechanism is model-agnostic. It can be integrated as a plug-in component to enhance the visual faithfulness of other powerful reasoning pipelines, suggesting a promising direction for future work in building more grounded and reliable VLMs.

\subsection{Ablation and Analysis}

\subsubsection{The Impact of Text Reasoning SFT}
We first evaluate the effectiveness of the Text Reasoning SFT stage in isolation to improve logical reasoning via language-only SFT. We fine-tune the base model on 20K logic-rich problems from the OpenThought dataset, which is distilled from the DeepSeek-R1 model. As shown in Table 4, this targeted SFT stage significantly boosts performance on the MathVista benchmark (from $57.78\%$ to $67.50\%$) without degrading performance on other tasks. This result confirms that language-only SFT is a highly effective method for improving a VLM's logical reasoning capabilities, providing a stronger logical foundation before we address visual perception.

\paragraph{Dense vs. Sparse: SFT Imitation vs. RL Correction for Perception}
Next, we investigate whether SFT can serve as a direct replacement for RL in teaching visual perception. To test this, we use our VLM judge model (GPT-4o) to distill solutions for the entire 80K RL training dataset in the ``See-Think-Answer'' format, and then supervised fine-tune the base model on this data. The results in Table~\ref{tab:mainresult} show that this SFT-only approach has limited success. While it improves the HallusionBench score in isolation, it fails to match the broad gains of our RL or hybrid methods. We hypothesize this is due to a reference-policy model distribution mismatch~\cite{sun2025warmup,taoyour}. SFT forces the policy model to perform token-level imitation of the reference's outputs, which may not align well with the student's own internal representations and capabilities. This has motivated us to train the perception stage through RL.

\subsubsection{The Impact of Perception Stage}
\label{sec:main_results}

Having improved the model's reasoning, we now focus on the perception stage. We study whether description rewards help beyond verifiable RL.
Table~\ref{tab:mainresult} presents our primary comparison between the \textbf{Verifiable RL} baseline (RL($\gamma=1$)) and our models that add a description reward ($r_{\mathrm{desc}}$) under either \emph{Aggregated} or \emph{Conditional} compositions. Across benchmarks, both \emph{Aggregated} and \emph{Conditional} consistently outperform Verifiable RL, with the largest gains on HallusionBench and visually intensive tasks (OCRBench, MathVista\_MINI). Notably, Verifiable RL underperforms the base model on HallusionBench (Table~\ref{tab:mainresult}), where its score drops from $43.80\%$ to $37.67\%$, indicating that answer-only supervision can inadvertently encourage hallucination-prone behaviors that score well during training but fail on explicit hallucination diagnostics. Please see examples in Appendix C. This confirms that adding direct VLM supervision to the perception stage is a crucial first step toward more faithful models.

\begin{table*}[]
\centering
\resizebox{0.70\linewidth}{!}{
\begin{tabular}{lccccc}
\toprule
\textbf{Model} & $\ge\textbf{ 1/8}$ & $\ge\textbf{ 2/8}$& $\ge\textbf{ 4/8}$ & $\textbf{8/8}$ & \textbf{Pass@1} \\
\midrule
\textbf{RL($\gamma=1$)} (Verifiable Rewards)                            & 0.850 & 0.779 & 0.678 & 0.408 & 0.6415 \\
\textbf{RL }(Aggregated Rewards)                       & 0.844 & 0.791 & 0.689 & 0.440 & 0.6545 \\
\textbf{RL($\gamma=0.5$)}  (Conditional Soft Gate)                & 0.848 & 0.784 & 0.696 & \textbf{0.443} & 0.6580 \\
\textbf{RL($\gamma=0$)} (Conditional Hard Gate)                & \textbf{0.864} & \textbf{0.808} & \textbf{0.705} & 0.420 & \textbf{0.6611} \\
\bottomrule
\end{tabular}
}
\caption{ Performance vs.\ consistency threshold (out of 8 samples) on MathVista\_MINI. Where $\ge k/8$ denotes the performance that the model scores at least $k$ correct responses out of $8$. The last column denotes Pass@1 accuracy over 8 samples.}
\label{tab:mathmvista}
\end{table*}

\begin{figure}[t]
    \centering
    \includegraphics[width=1.0\linewidth]{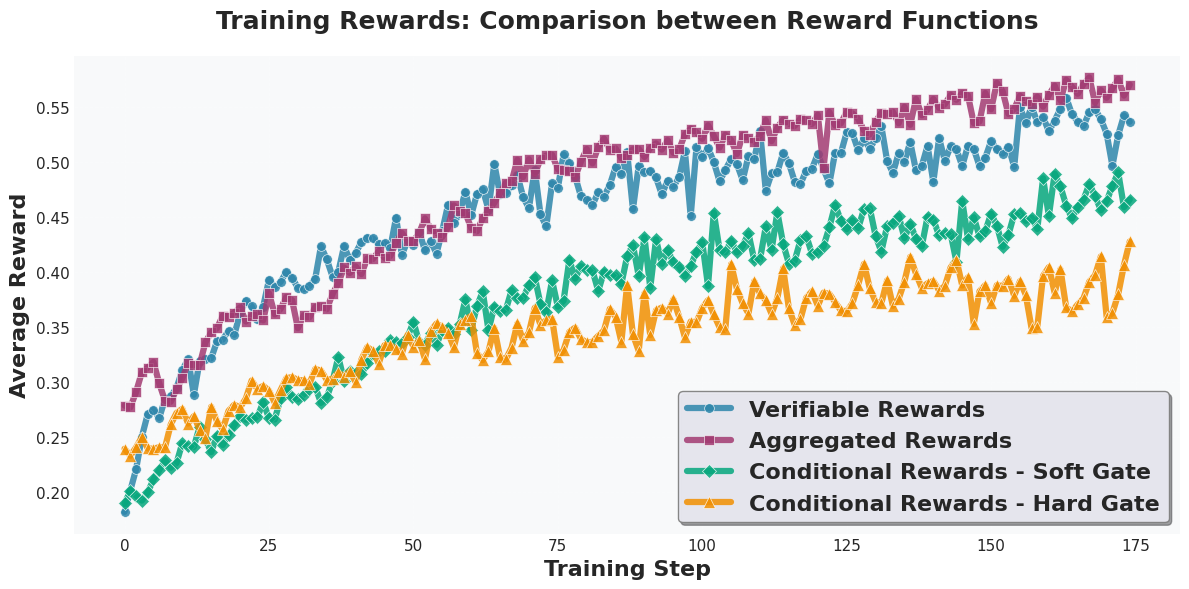}
    \caption{Training reward dynamics for each reward function. \emph{Aggregated} and Verifiable rise fastest; \emph{Conditional} grows more slowly.}
    \label{fig:rewards_train}
\end{figure}

\paragraph{Hard vs.\ soft gating.}
Beyond simply adding this signal, we find that the composition of the reward is critical for achieving the most robustly grounded reasoning. Our \emph{Conditional} reward models enforce a stricter causal link by making the answer reward contingent on a verified description. As the results in Table~\ref{tab:mainresult} show, this enforcement leads to the best overall performance. The \emph{Conditional} Hard Gate model, which uses a strict gate, achieves the highest scores across most benchmarks among RL models. It slightly outperforms the \emph{Conditional} Soft variant, especially on tasks sensitive to hallucination and math reasoning.

We hypothesize that this advantage comes from how different compositions handle "false positives"—rollouts that reach a correct final answer through a flawed or hallucinated reasoning process. While the \emph{Aggregated} model still grants partial credit in these cases, and the Soft-gated model offers a compromise, the Hard-gated model completely withholds the answer reward, thereby cutting off this shortcut. In the next section, we test this hypothesis by directly analyzing the correctness of the training rollouts.
\newline
\textbf{Takeaway.}
Supervising the perception stage with description reward not only improves accuracy on visually-intensive tasks like OCR and math but is essential for maintaining hallucination resilience. Our results show that a purely outcome-based Verifiable RL approach is insufficient and can inadvertently penalize model faithfulness, as evidenced by its underperformance on HallusionBench.

\begin{table*}[]
\centering
\resizebox{0.85\linewidth}{!}{
\begin{tabular}{lcccc}
\toprule
 & \multicolumn{2}{c}{\textbf{Final Answer Correct}} & \multicolumn{2}{c}{\textbf{Final Answer Incorrect}} \\
\cmidrule(lr){2-3}\cmidrule(lr){4-5}
\textbf{Model} & \;\textbf{Correct Roll}\; & \;\textbf{Wrong Roll}\; & \;\textbf{Correct Roll}\; & \;\textbf{Wrong Roll}\; \\
\midrule
\textbf{RL($\gamma=1$)} (Verifiable Rewards)          & 58.63 & 41.37 & 32.92 & 67.08 \\
\textbf{RL }(Aggregated Rewards)                     & 66.59 & 33.41 & 33.11 & 66.89 \\
\textbf{RL($\gamma=0.5$)}  (Conditional Soft Gate)               & 65.54 & 34.46 & 34.35 & 65.65 \\
\rowcolor{blue!10} \textbf{RL($\gamma=0$)} (Conditional Hard Gate)               & \textbf{66.77} & 33.23 & 32.41 & \textbf{67.59} \\
\bottomrule
\end{tabular}
}
\caption{Correctness of solution given whether the final answer is correct or not. Description supervision reduces false positives (correct answer, wrong solution).}
\label{tab:correctness}
\end{table*}

\subsubsection{Training Dynamics and Rollout Correctness }
\label{sec:dynamics}
\paragraph{Reward trajectories and sample efficiency.}

A direct look at the training reward curves in Figure~\ref{fig:rewards_train} shows that the Verifiable RL model appears to learn the fastest, with its average reward rising quickly in the initial training steps. The \emph{Aggregated} model follows a similar trajectory, also accumulating rewards efficiently due to the dense, combined signals, and even surpassing Verifiable RL in later steps. In contrast, the \emph{Conditional} variants learn much more slowly, as their reward is sparse, the final answer reward is frequently withheld until the description gate is passed. However, the test performance of the conditional models (as shown in Table~\ref{tab:mainresult}) indicates that the high training rewards of the Verifiable RL baseline are misleading. This suggests that the baseline might be engaging in reward hacking as it is optimizing for a flawed proxy (final answer correctness) by discovering shortcuts that do not rely on a grounded reasoning process.

\paragraph{Rollout audit: false positives vs.\ grounded solutions.}
To assess whether a high training reward corresponds to \emph{grounded} solutions, we extract \texttt{<think>} traces from training rollouts and ask the VLM to verify the correctness of the solution against the image and the question.

To verify this hypothesis, we conducted an audit of rollouts to assess whether the high training rewards correspond to genuinely correct solutions. We extracted the reasoning traces from the \texttt{<think>} traces of the training rollouts and used GPT-4o to verify their correctness against the image and question, independent of the final answer (we provide the prompts in Appendix C).

As shown in Table~\ref{tab:correctness}, \textbf{Verifiable RL} exhibits a high rate of \textbf{false positives}, cases where the final answer is correct yet the solution is wrong or ungrounded (\textbf{41.37\%} among correct-answer rollouts). This means that in over two-fifths of instances in which it received a positive reward for a correct answer, it did so despite an incorrect process. Introducing description supervision reduces this to \textbf{$\sim$33--35\%} across \emph{Aggregated}/\emph{Conditional} settings, with the \emph{Conditional} Hard model achieving the lowest rate at $33.23\%$. This reduction also mirrors the improvements on hallucination diagnostics in Table~\ref{tab:mainresult} and explains why description-supervised models can generalize better despite slower early reward growth.

\paragraph{Qualitative Analysis.}
Qualitative inspection shows some dominant failure modes under Verifiable RL, where the model invents details (e.g., colors, objects, textures) that are not present in the image to fit a reasoning template or even  contradicts the visual evidence. In other cases, the model provides a vague or incomplete description that omits a key visual element required for a correct logical step. We provide comparative examples in Appendix C.

\begin{table*}[!h]
\centering

\resizebox{0.85\linewidth}{!}{
\begin{tabular}{lccccc}
\toprule
\textbf{Model }                                             & \textbf{MathVista} & \textbf{MMMU} & \textbf{MMBench} & \textbf{OCRBench} & \textbf{Hallusion} \\
\midrule
\textbf{RL($\gamma=1$)} (Verifiable Rewards)                             & 64.15 & 51.42 & 83.20 & 85.12 & 37.67 \\
\textbf{RL($\gamma=0$)}  (\textbf{GPT-4o} Reward)               & \textbf{66.11} & \textbf{52.11} & \textbf{84.29} & \textbf{86.16} & \textbf{47.22} \\
\textbf{RL($\gamma=0$)}  (\textbf{Self} Reward)                  & 65.33 & 51.08 & 83.50 & 85.80 & 40.22 \\
\bottomrule
\end{tabular}
}
\caption{Ablation on VLM capacity (GPT-4o and Qwen2.5-VL-7B) for description reward (reporting Pass@1 accuracy).}
\label{tab:ablationjudge}
\end{table*}

\paragraph{Reliability and consistency under sampling.}

To assess whether reducing false positives leads to more stable outputs, we evaluate the model's reliability across multiple samples. We generate eight response candidates for each question and measure the success rate at varying levels of required consistency. The results on MathVista\_MINI in Table~\ref{tab:mathmvista} reveal a difference in robustness. To achieve a correct answer on at least four of the eight samples (a $50\%$ consistency threshold), the Verifiable RL model succeeds $67.8\%$ of the time, whereas the \emph{Conditional Hard} model succeeds $70.5\%$ of the time. This reliability gap persists and often widens at stricter thresholds (e.g., requiring all 8 samples to be correct), a trend also observed on other benchmarks (Appendix B). This shows that description-supervised models learn more fundamentally sound reasoning paths, making their correct answers more reproducible and less dependent on chance.

\medskip
\noindent\textbf{Summary.} The Verifiable RL model's faster reward growth might be misleading, as it is driven by rewarding a high volume of "false positive" solutions with ungrounded reasoning. By incorporating description supervision, particularly with conditional gating, we rectify this issue. This approach reduces the rate of false positives, which directly translates to greater resilience against hallucinations, and demonstrably more consistent and reliable outputs.

\subsubsection{Perception + Reasoning Synergy}

\label{sec:sft_rl}

Our PeRL-VL framework in Table~\ref{tab:mainresult} demonstrates a powerful synergy effect of both the text reasoning SFT stage and the perception stage. The final model preserves the substantial gains on MathVista from the SFT stage while improving on all other benchmarks, with a particularly dramatic increase on HallusionBench (from $41.40\%$ to $55.90\%$). This confirms the effectiveness of our decoupled approach: SFT sharpens logic and RL grounds that logic in visual reality.

\subsubsection{VLM Reward Model Ablation }

To investigate the impact of VLM model capability on supervision quality, we conduct an ablation study where we replace the GPT-4o model with the policy model itself (Qwen2.5-VL-7B) to provide the description reward, a form of self-supervision. As shown in Table~\ref{tab:ablationjudge}, this "Self Reward" model significantly outperforms the Verifiable RL baseline across all benchmarks. This result supports the hypothesis that verification is an easier task than generation. Even a moderately capable model can effectively identify and penalize many of its own ungrounded or hallucinated descriptions during on-policy training. However, the results also underscore the value of stronger supervision. The model supervised by GPT-4o remains consistently superior, achieving the highest performance across the board. The performance gap is most pronounced on HallusionBench, where the stronger judge's ability to detect subtle visual inaccuracies provides a critical advantage. This confirms that while self-supervision is a viable and efficient strategy, the judge's strength directly correlates with the quality of the resulting visual grounding. For practitioners using the self-reward approach, we recommend periodically updating the reward model with the latest policy weights to ensure supervision quality improves in tandem with the student's capabilities.

\subsection{Limitations and Practical Guidance}
While our framework improves visual extraction and reasoning, its mandatory description step can add overhead, suggesting future work in efficiency description reward. For practitioners, a brief SFT cold or warm start can ensure improved reasoning or format adoption. Further, one may consider a possible gating curriculum (soft to hard) for the reward function, to stabilize early training, but minimizing ``false positives.'' Lastly, to balance tradeoffs, a hybrid judge schedule for reward function might improve efficiency.

\section{Conclusion}

In this work, we addressed the limitations of the standard RLVR paradigm when applied to vision-language models. We identified that its outcome-only nature leads to critical failure modes, encouraging "false positive" shortcuts where models are rewarded for correct answers derived from unfaithful reasoning. Our solution, PeRL-VL, is a decoupled framework that targets these issues separately and effectively.
For perception, we moved beyond simply adding a description reward to systematically explore different reward designs. Our investigation revealed that the structure of the reward function is a critical factor: designs that enforce a stronger causal link between a faithful description and a correct final answer are demonstrably superior. This approach directly mitigates reward hacking and leads to state-of-the-art performance on hallucination benchmarks. For reasoning, we showed that a dedicated Text Reasoning SFT stage effectively improves logical consistency and synergizes powerfully with our perception-focused RL, allowing the model to achieve strong performance across diverse tasks.
By explicitly and separately supervising the intermediate processes of perception and reasoning, and by demonstrating the critical importance of reward design, our work offers a principled path toward building VLMs that are not only more accurate but also more faithful and reliable.

\section{Acknowledgment}
Ruoxi Jia and the ReDS lab acknowledge support through grants from the National Science Foundation under grants IIS-2312794, IIS-2313130, and OAC-2239622.

\clearpage

{
    \small
    \bibliographystyle{plainnat}
    \bibliography{references.bib}
}

\clearpage
\setcounter{page}{1}
\onecolumn


  \appendix
\section*{Appendix}
\label{sec:appendix}

\section{Appendix A: Experimental Settings and Implementation Details}
\label{app:details}

\subsection{Dataset Details}
\paragraph{Training Data.}
Our primary training dataset consists of a curated \textbf{80K sample} mixture. This mixture is aggregated from two high-quality open-source datasets:
\begin{itemize}
    \item \textbf{ThinkLite-VL}~\cite{wang2025sotawithless}: Focused on finding difficult (requiring multiple steps to reach a solution) and solvable visual reasoning problems.
    \item \textbf{VL-Cogito}~\cite{yuan2025vlcogito}: Focused on both difficulty (solved less than 50\% of times) and the coverage of the problems. Additionally, enforcing open-ended format of the problems.

\end{itemize}
We combined these datasets resulting in a $\sim$ 80K dataset, after some filtering with deduplication. For the \textbf{Text Reasoning SFT} stage, we used a \textbf{20K sample} subset of the \textbf{OpenThought}~\cite{guha2025openthoughts} dataset, specifically filtering for math and coding problems distilled from DeepSeek-R1 to maximize logical reasoning.

\paragraph{Evaluation Benchmarks.}
We evaluate on the standard splits of the following benchmarks available in VLMEvalKit:
\begin{itemize}
    \item \textbf{OCRBench}~\cite{liu2024ocrbench}: Assesses text recognition within images.
        \begin{itemize}
        \item 1000 points, accuracy on all points
        \end{itemize}
    \item \textbf{MathVista\_MINI}~\cite{lu2023mathvista}: A subset of MathVista focused on visual mathematics (geometry, charts, functions).
        \begin{itemize}
        \item 1000 points, accuracy on all points
        \end{itemize}
    \item \textbf{MMMU\_DEV\_VAL}~\cite{yue2024mmmu}: Multidisciplinary questions requiring expert-level knowledge and reasoning.
        \begin{itemize}
        \item 1050 points, accuracy on all points, including development and validation points
        \end{itemize}
    \item \textbf{MMBench\_DEV\_EN\_V11}~\cite{liu2024mmbench}: A comprehensive multiple-choice benchmark for general visual capabilities.
        \begin{itemize}
        \item 4876 points, accuracy on all points
        \end{itemize}
    \item \textbf{HallusionBench}~\cite{guan2024hallusionbench}: A benchmark designed to detect visual hallucinations and consistency errors.
        \begin{itemize}
        \item 951 points, accuracy on all points
        \end{itemize}
\end{itemize}

\subsection{Model Architecture}
\begin{itemize}
    \item \textbf{Policy Model ($\pi_\theta$):} We use \textbf{Qwen2.5-VL-7B-Instruct} and \textbf{Qwen2.5-VL-3B-Instruct} for our experiments. 
    \item \textbf{Reward Model ($\mathcal{T}$):} We use \textbf{GPT-4o}  via API for providing the description reward ($r_{\mathrm{desc}}$) and for measuring correctness of the logical steps in \texttt{<think>} traces. For the ablation study in Section 4.5, we use the policy model itself as the judge. We also use \textbf{GPT-4o} to generate training data for SFT used in Section~\ref{sec:sft}.
    \item \textbf{Baseline Models:}  We benchmark against several strong, publicly available baseline models with our evaluationg code: ThinkLite‑VL~\cite{wang2025sotawithless}, VL‑Cogito~\cite{yuan2025vlcogito}, R1-Share‑VL‑7B~\cite{yao2025r1}, PeBR‑R1‑7B~\cite{chen2025perception}, Vision-SR1~\cite{li2025self}. 
\end{itemize}

\subsection{Training Hyperparameters}
All experiments were conducted on a cluster of $8 \times$ NVIDIA A100 (80GB) GPUs.

\paragraph{Stage 1: SFT.}
For supervsised fine-tuning, we use the LLaMA-Factory~\cite{zheng2024llamafactory} framework with the following settings in Table~\ref{tab:sft_hyper}.
\begin{table*}[h!]
\centering
\resizebox{0.75\linewidth}{!}{
\begin{tabular}{ll}
\toprule
\textbf{Hyperparameter} & \textbf{Value} \\
\midrule
Finetuning Type & Full \\
Tower Frozen & Vision \\
Global Batch Size & 256 \\
Batch Size Per Device & 1 \\
Gradient Accumulation Steps & 32 \\
Learning Rate & $5 \times 10^{-6}$ \\
Epochs & 5 \\
LR Scheduler & Cosine \\
Warmup Ratio & 0.1 \\
Max Sequence Length & 32,768 \\
Optimizer & AdamW ($\beta_1=0.9, \beta_2=0.95$) \\
\bottomrule
\end{tabular}
}
\caption{Hyperparameters for SFT Stage.}
\label{tab:sft_hyper}
\end{table*}

\paragraph{Stage 2: RL (GRPO).}
We use the VERL library~\cite{sheng2024hybridflow} for RL training with Group-Relative Policy Optimization using the hyperparameters in Table~\ref{tab:rl_hyper}.

\begin{table*}[h!]
\centering
\resizebox{0.80\linewidth}{!}{
\begin{tabular}{ll}
\toprule
\textbf{Hyperparameter} & \textbf{Value} \\
\midrule
Global Batch Size & 512 \\
PPO Mini Batch Size & 128 \\
Rollouts per Input ($K$) & 16 \\
Learning Rate & $1 \times 10^{-6}$ \\
Epochs  & 1 \\
Max Sequence Length & 20,480 \\
KL Coefficient ($\beta$) & 0.001 \\
Entropy Coefficient ($\beta$) & 0.001 \\
Reward Weights ($\alpha_{\mathrm{fmt}}, \alpha_{\mathrm{desc}}, \alpha_{\mathrm{ans}}$) & $0.1, 0.45/0, 0.45/0.9$ \\
Gating Parameter ($\gamma$) & $0$ (Hard), $0.5$ (Soft), $1.0$ (Baseline) \\
\bottomrule
\end{tabular}
}
\caption{Hyperparameters for RL Stage.}
\label{tab:rl_hyper}
\end{table*}

\paragraph{Evaluation.}
For benchmarking datasets, we use VLMEvalKit~\cite{duan2024vlmevalkit} datasets with vLLM~\cite{kwon2023efficient} for efficient inference with hyperparameters in Table~\ref{tab:eval_hyper} and \textbf{Qwen2.5-72B-Instruct} for answer evaluation against the ground truth.

\begin{table*}[h!]
\centering
\resizebox{0.40\linewidth}{!}{
\begin{tabular}{ll}
\toprule
\textbf{Hyperparameter} & \textbf{Value} \\
\midrule
GPU Memory Utilitzation & 0.9 \\
DType & bfloat16 \\
Max Model Length & 32000 \\
Temperature & 0.6 \\
Number of Samples & 8 \\
Repetition Penalty & 1.0 \\

\bottomrule
\end{tabular}
}
\caption{Hyperparameters for inference.}
\label{tab:eval_hyper}
\end{table*}

\newpage

\section{Appendix B: Additional Experimental Results}
\label{app:results}

\subsection{Full Consistency Analysis}

In the main text, we analyzed model output reliability using multi-sample consistency metrics for MathVista dataset. Here, we extend this analysis to MMMU\_DEV\_VAL (Table~\ref{tab:mmmu}), OCRBench (Table~\ref{tab:ocrbench}), MMBench\_DEV\_EN\_V11 (Table~\ref{tab:mmbench}), and HallusionBench (Table~\ref{tab:hallusion}). These additional results confirm that the stability improvements driven by RL with perception reward are robust and consistent across diverse benchmarks.

\begin{table*}[!ht]
\centering
\begin{tabular}{lccccc}
\toprule
\textbf{Model} & $\ge\textbf{ 1/8}$ & $\ge\textbf{ 2/8}$& $\ge\textbf{ 4/8}$ & $\textbf{8/8}$ & \textbf{Pass@1} \\
\midrule
\textbf{RL($\gamma=1$)} (Verifiable Rewards)                         & 0.943 & 0.915 & 0.869 & 0.726 & 0.8513 \\
\textbf{RL }(Aggregated Rewards)                       & 0.940 & \textbf{0.925} & 0.882 & 0.758 & 0.8630 \\
\textbf{RL($\gamma=0.5$)}  (Conditional Soft Gate)                & 0.942 & 0.921 & 0.882 & \textbf{0.764} & \textbf{0.8656} \\
\textbf{RL($\gamma=0$)} (Conditional Hard Gate)               & \textbf{0.945} & 0.924 & \textbf{0.882} & 0.742 & 0.8616 \\
\bottomrule
\end{tabular}
\caption{Performance vs.\ consistency threshold (out of 8 samples) on OCRBench. Where $\ge k/8$ denotes the performance that the model scores at least $k$ correct responses out of $8$. The last column denotes Pass@1 accuracy over 8 samples.}
\label{tab:ocrbench}
\end{table*}

\begin{table*}[!ht]
\centering
\begin{tabular}{lccccc}
\toprule
\textbf{Model} & $\ge\textbf{ 1/8}$ & $\ge\textbf{ 2/8}$& $\ge\textbf{ 4/8}$ & $\textbf{8/8}$ & \textbf{Pass@1} \\
\midrule
\textbf{RL($\gamma=1$)} (Verifiable Rewards)                           & 0.775 & \textbf{0.691} & 0.541 & 0.254 & 0.5142 \\
\textbf{RL }(Aggregated Rewards)                       & \textbf{0.778} & 0.673 & 0.535 & 0.257 & 0.5080 \\
\textbf{RL($\gamma=0.5$)}  (Conditional Soft Gate)               & 0.750 & 0.657 & \textbf{0.551} & \textbf{0.298} & 0.5209 \\
\textbf{RL($\gamma=0$)} (Conditional Hard Gate)                 & 0.756 & 0.671 & 0.550 & 0.289 & \textbf{0.5210} \\
\bottomrule
\end{tabular}
\caption{Performance vs.\ consistency threshold (out of 8 samples) on MMMU\_DEV\_VAL. Where $\ge k/8$ denotes the performance that the model scores at least $k$ correct responses out of $8$. The last column denotes Pass@1 accuracy over 8 samples.}
\label{tab:mmmu}
\end{table*}

\begin{table*}[!ht]
\centering
\begin{tabular}{lccccc}
\toprule
\textbf{Model} & $\ge\textbf{ 1/8}$ & $\ge\textbf{ 2/8}$& $\ge\textbf{ 4/8}$ & $\textbf{8/8}$ & \textbf{Pass@1} \\
\midrule
\textbf{RL($\gamma=1$)} (Verifiable Rewards)                           & \textbf{0.951} & \textbf{0.920} & 0.856 & 0.674 & 0.8320 \\
\textbf{RL }(Aggregated Rewards)                      & \textbf{0.951} & 0.918 & 0.859 & 0.672 & 0.8320 \\
\textbf{RL($\gamma=0.5$)}  (Conditional Soft Gate)                 & 0.942 & 0.913 & \textbf{0.861} & 0.710 & 0.8415 \\
\textbf{RL($\gamma=0$)} (Conditional Hard Gate)                 & 0.938 & 0.912 & \textbf{0.861} & \textbf{0.722} & \textbf{0.8429} \\
\bottomrule
\end{tabular}
\caption{Performance vs.\ consistency threshold (out of 8 samples) on MMBench\_DEV\_EN\_V1.1. Where $\ge k/8$ denotes the performance that the model scores at least $k$ correct responses out of $8$. The last column denotes Pass@1 accuracy over 8 samples.}
\label{tab:mmbench}
\end{table*}

\begin{table*}[!ht]
\centering
\begin{tabular}{lccccc}
\toprule
\textbf{Model} & $\ge\textbf{ 1/8}$ & $\ge\textbf{ 2/8}$& $\ge\textbf{ 4/8}$ & $\textbf{8/8}$ & \textbf{Pass@1} \\
\midrule
\textbf{RL($\gamma=1$)} (Verifiable Rewards)                           & 0.582 & 0.489 & 0.394 & 0.196 & 0.3767 \\
\textbf{RL }(Aggregated Rewards)                        & 0.658 & 0.576 & \textbf{0.496} & \textbf{0.288} & \textbf{0.4730} \\
\textbf{RL($\gamma=0.5$)}  (Conditional Soft Gate)                & \textbf{0.689} & \textbf{0.621} & 0.481 & 0.219 & 0.4610 \\
\textbf{RL($\gamma=0$)} (Conditional Hard Gate)                 & 0.687 & 0.594 & 0.492 & 0.281 & 0.4720 \\
\bottomrule
\end{tabular}
\caption{Performance vs.\ consistency threshold (out of 8 samples) on HallusionBench. Where $\ge k/8$ denotes the performance that the model scores at least $k$ correct responses out of $8$. The last column denotes Pass@1 accuracy over 8 samples.}
\label{tab:hallusion}
\end{table*}

\subsection{Results on ThinkLite-VL.}

We present results for models trained exclusively on the ThinkLite-VL dataset in Table~\ref{tab:thinklite}. Consistent with our main findings, RL with description rewards continues to outperform the baseline (Verifiable Rewards) even on this smaller dataset. However, the magnitude of these gains is reduced compared to models trained on our larger, aggregated dataset (ThinkLite-VL + VL-Cogito). This suggests that while our method is effective on smaller datasets, its full potential is realized when applied to larger data sources.

\begin{table*}[!ht]
\centering

\resizebox{0.999\linewidth}{!}{
\begin{tabular}{llccccc}
\toprule
\textbf{Dataset }   & \textbf{Model }                                             & \textbf{MathVista} & \textbf{MMMU} & \textbf{MMBench} & \textbf{OCRBench} & \textbf{Hallusion} \\
\midrule
ThinkLite-VL dataset   & \textbf{RL($\gamma=1$)} (Verifiable Rewards)                           & 62.08 & 50.23 & 83.20 &  \textbf{86.10} & 35.54 \\
11K  & \textbf{RL($\gamma=0$)}  (\textbf{GPT-4o} Reward)               & \textbf{62.96} & \textbf{50.45} & \textbf{83.52} & 85.73 & \textbf{43.84} \\
\midrule
New dataset  & \textbf{RL($\gamma=1$)} (Verifiable Rewards)                              & 64.15 & 51.42 & 83.20 & 85.12 & 37.67 \\
 80K  & \textbf{RL($\gamma=0$)}  (\textbf{GPT-4o} Reward)               & \textbf{66.11} & \textbf{52.11} & \textbf{84.29} & \textbf{86.16} & \textbf{47.22} \\
\bottomrule
\end{tabular}
}
\caption{Performance comparison between two datasets, ThinkLite-VL and the new composed dataset (reporting Pass@1 accuracy).}
\label{tab:thinklite}
\end{table*}

\subsection{Results on different benchmarks.}

Here, we provide results on additional benchmarks, MMVet~\cite{yu2023mm}, MMStar~\cite{chen2024we}, MathVision\_MINI~\cite{wang2024measuring}, RealWorldQA~\cite{realworldqa}, and DynaMath~\cite{zou2025dynamath}, for more comprehensive results. We observe in Table~\ref{tab:add_bench} that training RL with description reward can improve upon training with the verifiable rewards only (RL($\gamma=1$)). Additionally, training with reasoning SFT can further boost performance of the model. PeRL-VL reaches competitive performance with baseline models.

To provide a more comprehensive evaluation, we report results on five additional benchmarks: MMVet\cite{yu2023mm}, MMStar\cite{chen2024we}, MathVision\_MINI\cite{wang2024measuring}, RealWorldQA\cite{realworldqa}, and DynaMath\cite{zou2025dynamath}. As shown in Table\ref{tab:add_bench}, RL training with description rewards consistently outperforms the baseline trained with verifiable rewards only (RL($\gamma=1$)). Furthermore, incorporating reasoning SFT yields additional performance gains, enabling PeRL-VL to demonstrate competitive performance against strong baselines across this expanded suite of tasks.

\begin{table*}[!ht]
\centering
\resizebox{0.99\textwidth}{!}{
\begin{tabular}{lcccccc}

\toprule
                                              & \textbf{DynaMath} & \textbf{MMVet} & \textbf{MMStar} & \textbf{RealWorldQA} & \textbf{MathVision\_MINI} & \textbf{AVG}    \\
                                              \toprule
Base (Qwen2.5-VL-7B-Instruct)                            & 43.01    & 67.45 & 52.67  & 55.49       & 24.42            & 48.608 \\
RL($\gamma=1$) (Verifiable Rewards)                & 49.62    & 70.87 & 58.12  & 62.82       & 27.01            & 53.688 \\
\textbf{RL($\gamma=0$) } (Conditional Hard Gate)      & 50.52    & 71.27 & 58.56  & 63.80       & 27.71            & 54.372 \\ \rowcolor{blue!10}
\textbf{PeRL-VL} & 51.01    & 72.22 & 59.95  & 64.72       & 30.98            & 55.776 \\
\midrule
VL Cogito                                     & 50.42    & 69.15 & 54.79  & 56.01       & 31.00            & 52.274 \\
ThinkLite7B                                   & 49.70    & 69.72 & 56.73  & 61.73       & 27.59            & 53.094 \\
R1ShareVL7B                                   & 45.23    & 69.61 & 57.43  & 63.33       & 26.69            & 52.458 \\
PeBR-R1-7B                                    & 47.48    & 74.94 & 60.30  & 64.67       & 33.10            & 56.098 \\
SelfRewarded-R1-7B                            & 41.94    & 71.10 & 59.3   & 66.96       & 27.79            & 53.418 \\
\bottomrule
\end{tabular}
}
\caption{Performance on additional benchmarks for models trained with different reward functions or methods, including the base model and SFT trained models. Comparison with open-source baselines. All results are Pass@1.}
\label{tab:add_bench}
\end{table*}

\subsection{Results on Qwen2.5-VL-3B-Instruct.}

To demonstrate the generalizability of our approach, we evaluate PeRL-VL on a smaller backbone, Qwen2.5-VL-3B-Instruct. The results, presented in Table~\ref{tab:3b_newbench} and Table~\ref{tab:3b_bench}, confirm that our method remains effective across different model scales.

\begin{table*}[!ht]
\resizebox{0.99\textwidth}{!}{
\begin{tabular}{lcccccc}
\toprule
                                     & \textbf{MathVista} & \textbf{MMMU} & \textbf{MMBench} & \textbf{OCRBench} & \textbf{Hallusion} & \textbf{AVG} \\
                                     \toprule
Base (Qwen2.5-VL-3B-Instruct)          & 42.70              & 36.74         & 64.56            & 61.56             & 48.00              & 50.712       \\
RL($\gamma=1$) (Verifiable Rewards)        & 51.31              & 47.11         & 75.64            & 78.20             & 28.32              & 56.116       \\
\textbf{RL($\gamma=0$) } (Conditional Hard Gate)       & 53.96              & 46.69         & 76.31            & 77.89             & 51.82              & 61.334       \\
\textbf{PeRL-VL-3B} & 56.12              & 47.34         & 76.50            & 78.12             & 53.48              & 62.312       \\
\bottomrule
\end{tabular}
}
\caption{Performance on benchmarks for Qwen2.5-VL-3B-Instruct trained with different reward functions. All results are Pass@1.}
\label{tab:3b_bench}
\end{table*}

\begin{table*}[!ht]
\resizebox{0.99\textwidth}{!}{
\begin{tabular}{lcccccc}
\toprule
                                     & \textbf{DynaMath} & \textbf{MMVet} & \textbf{MMStar} & \textbf{RealWorldQA} & \textbf{MathVision\_MINI} & \textbf{AVG}    \\
                                     \toprule
Base (Qwen2.5-VL-3B-Instruct)          & 32.16    & 50.69 & 37.95  & 45.39       & 18.71            & 36.98  \\
RL($\gamma=1$) (Verifiable Rewards)        & 41.44    & 63.10 & 43.33  & 54.10       & 23.97            & 45.188 \\
\textbf{RL($\gamma=0$) } (Conditional Hard Gate)         & 41.72    & 62.96 & 44.73  & 54.73       & 23.68            & 45.564 \\
\textbf{PeRL-VL-3B} & 42.39    & 63.59 & 45.35  & 56.43       & 24.59            & 46.47  \\
\bottomrule
\end{tabular}
}
\caption{Performance on additional benchmarks for Qwen2.5-VL-3B-Instruct trained with different reward functions. All results are Pass@1.}
\label{tab:3b_newbench}
\end{table*}

\subsection{Reward Prompts}
To compute the description reward ($r_{\mathrm{desc}}$), we use the following prompt template with VLM reward model. The prompt is designed to check for both faithfulness (hallucination) and sufficiency.

\begin{tcolorbox}[title=Description Verification Prompt, colback=gray!5, colframe=black!75, width=\linewidth]
\textbf{System:} You are a helpful assistant that checks correctness, detailness, and completeness of each sentence in the description to the image. You also check if the description is correct and provides sufficient information to answer the provided question.

\textbf{Instructions:}
Given the image, check if the provided description (for each sentence) is correct, unambiguous and would the description as a whole be sufficient to solve the following problem even without seeing the image. \\
Problem:\{problem\}
\\
Here is the description for you to check: \{description\}

To check correctness, for each sentence, provide which sentence you are checking in \textbackslash sentence\{<SENTENCE>\}, your explanation for the score in \textbackslash reasoning\{<YOUR REASONING>\} and the score in \textbackslash score\{<SCORE>\} for each sentence.
Then at the end, if all the sentences are correct, unambiguous, and fully sufficient to solve the problem, then output the outcome 1 in the \textbackslash outcome\{<OUTCOME>\} format, \textbackslash outcome\{1\}.
Otherwise, output the outcome 0 in the \textbackslash outcome\{<OUTCOME>\} format, \textbackslash outcome\{0\}. 
\\
Example output format (numbers are placeholders): \textbackslash sentence\{...\}
\textbackslash reasoning\{...\}
\textbackslash score\{0\}
...
The description is insufficient to solve the problem, because ... . Thus, the outcome is \textbackslash outcome\{0\}.
\end{tcolorbox}

To compute the correctness of the $\text{<think>}$ traces, we use the following prompt template with GPT-4o. The prompt is designed to check the logical correctness and faithfulness of each step.

\begin{tcolorbox}[title=Think Verification Prompt, colback=gray!5, colframe=black!75, width=\linewidth]
\textbf{System:} You are a helpful assistant that checks correctness of each logical step in the solution to a problem. You also check if the step uses correct information from the image.

\textbf{Instructions:}
Given the image, check if the provided solution to a problem is correct and does not have any wrong logical steps or wrong references to the image.\\
Image:{image} \\
 
Problem:{problem} \\

Ground Truth: \{groundtruth\} \\

Here is the solution for you to check: \{solution\}\\

Carefully analyze the solution by breaking down into individual step. For each step in the solution, if you find a step that is logically wrong or makes an incorrect reference to the image, please output \textbackslash OUTCOME\{0\}. Otherwise, output \textbackslash OUTCOME\{1\} if all steps are correct.
\end{tcolorbox}

\newpage

To generate SFT training data from GPT-4o for Section~\ref{sec:main_results}, we use the following prompt:

\begin{tcolorbox}[title=Generation Prompt, colback=gray!5, colframe=black!75, width=\linewidth]
\textbf{System:} Please think step by step and check carefully all details in the image.\\
   Before solving the problem, You FIRST think about the description of the image as an internal monologue and then provide the description. Make sure the description contains necessary details to answer the given problem.\\
  The description process MUST BE enclosed within $<\text{description}>$ $</\text{description}>$ tags. \\
   Then proceed to solving the problem step by step and provide your thought process between <think $</\text{think}>$ tags. Lastly, provide the summary of your solution between the $<\text{answer}>$ $</\text{answer}>$ tags with the key result enclosed within \textbackslash boxed\{\}.\\

\textbf{Instructions:}
Given a question and an image, please solve the problem.\\
Question: \{Question\} \\
Image: \{Image\} \\

The ground truth should be: \{groundtruth\}.
\end{tcolorbox}

To measure the correctness of the final answer to the groundtruth answer of the benchmark, we use a capable language model, Qwen2.5-72B-Instruct, to match the answer instead of only relying on the parsing, which is often inflexible. We use the following prompt for evaluation.

\begin{tcolorbox}[title=Evaluation Prompt, colback=gray!5, colframe=black!75, width=\linewidth]
\textbf{System:} You are a helpful assistant. Your goal is to extract my answer letter from my response.

\textbf{Instructions:}
Given the question problem: \{problem\}.\\
Z. no previous option matches my response\\

Given my response: \\
<START OF MY RESPONSE>\\
$\{\text{MY RESPONSE}\}$.\\
<END OF MY RESPONSE>\\

Do not solve the problem. First extract the final answer from MY RESPONSE, then write down explicitly again the above answer choices, and lastly Check which of the options my response best matches.
If MY RESPONSE has not reached a clear conclusion, then choose other option.
Ignore the upper and lower cases.
Return the final answer and place in \textbackslash extracted\{\}, and the above answer choice in \textbackslash choices\{\} and the answer letter that best matches my final answer in \textbackslash letter\{\} format.
\end{tcolorbox}

\subsection{Qualitative Comparison: Baseline vs. PeRL-VL}

We provide representative examples where the \textbf{Verifiable RL} baseline produces false positives (Correct Answer, Wrong Reasoning), wrong visual extractions, wrong logical reasoning, or inconsistent reasoning to match the final answer.

vague or incomplete description that omits a key visual element required for a correct logical step

We provide representative examples from the test set where the \textbf{Verifiable RL} baseline produces a "False Positive" (Correct Answer, Wrong Reasoning) and how \textbf{PeRL-VL} corrects it.

\begin{figure*}[!ht]
\begin{MyBox}[tabularx*={\arrayrulewidth0.5mm}{X|X},title=Question: Is the number of tiny objects that are behind the small metal jet less than the number of tiny things left of the tiny sedan?
Choices:
(A) Yes
(B) No. Answer: A \thetcbcounter, label=def:a, fontupper=\small]%

     \multicolumn{2}{p{0.96\linewidth}}{\centering \includegraphics[width=0.6\linewidth]{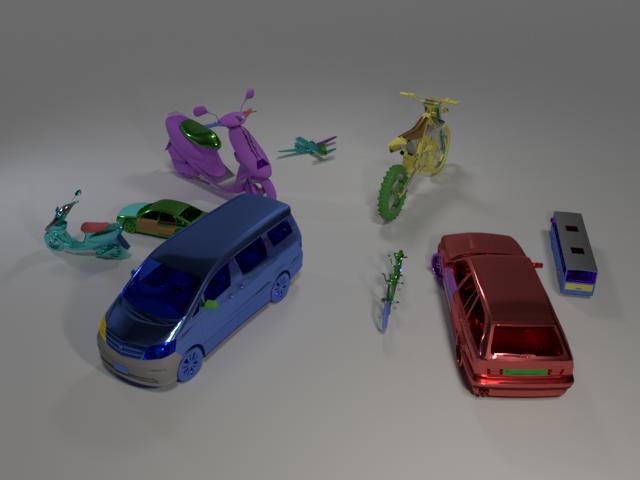}} 
    \\ \hline
\textbf{RL with Verifiable Rewards:} <think>
To answer this question, let's break it down step by step:

1. Identify the "small metal jet": In the image, there is a small metallic object that appears to be a jet, located towards the back of the scene.
2. Count the "tiny objects behind the small metal jet": There are no tiny objects directly behind the small metal jet. The objects behind it are a purple scooter and a green car.
3. Identify the "tiny sedan": In the image, \textbf{\textcolor{red}{there is a small sedan, which is a red car.}}
4. Count the "tiny things left of the tiny sedan": To the left of the tiny sedan, there is a green scooter and a green car.

Now, compare the two quantities:
- The number of tiny objects behind the small metal jet is 0.
- The number of tiny things left of the tiny sedan is 2.

Since 0 is less than 2, the number of tiny objects behind the small metal jet is indeed less than the number of tiny things left of the tiny sedan.

Therefore, the answer is Yes.
</think>
<answer>
\textbackslash \textbf{\textcolor{ForestGreen}{boxed\{A\} }}
</answer>
    
    &
\textbf{PeRL-VL: } <description>
The image contains various objects, including vehicles and other items. The objects are positioned on a flat surface. Here is a list of the objects and their approximate positions:

1. A small metal jet is located towards the back of the image.
2. A tiny sedan is positioned to the left of the image.
3. There are several other vehicles and objects scattered around the image, including a large blue van, a red wagon, a purple scooter, a yellow dirt bike, and a small blue bus.

The question asks about the number of tiny objects behind the small metal jet versus the number of tiny objects to the left of the tiny sedan. We need to count the relevant objects for both comparisons.
</description>

<think>
To answer the question, we need to identify and count the tiny objects behind the small metal jet and the tiny objects to the left of the tiny sedan.

1. **Counting tiny objects behind the small metal jet:**
   - There are no tiny objects directly behind the small metal jet. 

2. **Counting tiny objects to the left of the tiny sedan:**
   - The tiny sedan is located towards the \textbf{\textcolor{ForestGreen}{left side of the image. To the left of the tiny sedan, there is a small metal scooter}}. This is the only tiny object to the left of the tiny sedan.

Based on the above analysis, the number of tiny objects behind the small metal jet is 0, and the number of tiny objects to the left of the tiny sedan is 1. Therefore, the number of tiny objects behind the small metal jet is less than the number of tiny objects to the left of the tiny sedan.
</think>
<answer>
The total volume of the measuring cup is \textbackslash boxed\{A\}.
</answer>
\end{MyBox}
\end{figure*}


\begin{figure*}[!ht]
\begin{MyBox}[tabularx*={\arrayrulewidth0.5mm}{X|X},title=Question: what is the total volume of the measuring cup? (Unit: g).  Answer: 1000 \thetcbcounter, label=def:b, fontupper=\small]%

     \multicolumn{2}{p{0.96\linewidth}}{\centering \includegraphics[width=0.6\linewidth]{fig/dat2_mv.jpg}} 
    \\ \hline
   \textbf{RL with Verifiable Rewards:}  <think>
The question asks for the total volume of the measuring cup. The measuring cup is marked with volume measurements in milliliters (ml). \textbf{\textcolor{red}{The cup appears to be filled with a substance, likely flour, up to the 500 ml mark. Therefore, the total volume of the measuring cup is 500 ml.}}

Since the question asks for the answer in grams (g), we need to consider the density of the substance in the cup. However, the question does not provide the density of the substance, and typically, the volume is asked in its natural unit without conversion unless specified. Given the context and the typical use of measuring cups, the volume is usually given in the same unit as the markings on the cup.

Thus, the total volume of the measuring cup is 500 ml, which is equivalent to 500 grams if the substance is at standard density (1 g/ml for common substances like flour).
</think>
<answer>
\textbackslash \textbf{\textcolor{red}{boxed\{500\} }}
</answer>
    
    &
    {
    \textbf{PeRL-VL: } <description>
The image shows a kitchen setup with a measuring cup on the right side. The measuring cup is labeled with measurements in liters and grams, indicating it is used for measuring ingredients, such as sugar, flour, or other dry goods. The cup is partially filled with a substance that appears to be flour. The background includes a bowl of flour, an egg carton with eggs, and a wooden board. The measuring cup has a scale that ranges from 0 to 1000 grams, with corresponding measurements in liters. The question asks for the total volume of the measuring cup, which is a measurement in grams.
$</\text{description}>$

<think>
The measuring cup in the image is marked with a scale that measures in grams and liters. The question asks for the total volume of the measuring cup, which is the maximum measurement it can hold. \textbf{\textcolor{ForestGreen}{From the image, the maximum measurement marked on the cup is 1000 grams}}. This is the total volume of the measuring cup, as it is a measure of capacity and the question is asking for the total volume it can hold.
</{think}>

<answer>
The total volume of the measuring cup is \textbackslash \textbf{\textcolor{ForestGreen}{boxed\{1000\}.}}
</{answer}>
}
\end{MyBox}
\end{figure*}

\begin{figure*}[!ht]
\begin{MyBox}[tabularx*={\arrayrulewidth0.5mm}{X|X},title=Question: Each of the following situations relates to a different company. <image 1> For company B\text{,} find the missing amounts.
A. $\$63\text{,}020$
B. $\$58\text{,}410$
C. $\$71\text{,}320$
D. $\$77\text{,}490$. Answer: D \thetcbcounter, label=def:c, fontupper=\small]%

     \multicolumn{2}{p{0.96\linewidth}}{\centering \includegraphics[width=0.6\linewidth]{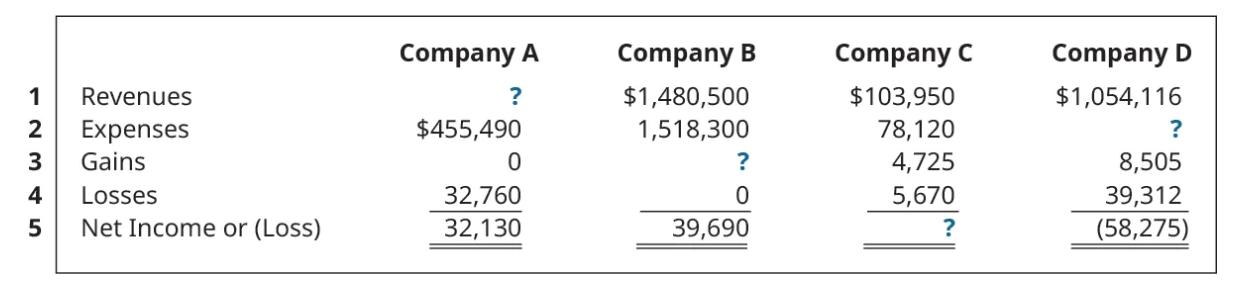}} 
    \\ \hline
    \textbf{RL with Verifiable Rewards:} <think>
To find the missing amounts for Company B, we need to use the information provided in the table to calculate the missing values. Let's start with the missing revenue for Company B.

1. **Revenue for Company B:**
   - We know the expenses for Company B are $\$1,518,300$.
   - We know the net income for Company B is $\$39,690$.
   - We can use the formula: Net Income = Revenue - Expenses.
   - Rearranging the formula to find Revenue: Revenue = Net Income + Expenses.
\textbf{\textcolor{red}{   - Plugging in the values: Revenue = $\$39,690$ + $\$1,518,300$ = $\$1,557,990$.}}

2. **Gains for Company B:**
   - We know the losses for Company B are $\$0$.
   - We know the net income for Company B is $\$39,690$.
   - We can use the formula: Net Income = Revenue - Expenses + Gains - Losses.
   - Rearranging the formula to find Gains: Gains = Net Income + Losses - Revenue + Expenses.
   - Plugging in the values: Gains = $\$39,690 + \$0 - \$1,557,990 + \$1,518,300 = \$0$.

3. **Expenses for Company D:**
   - We know the revenue for Company D is $\$1,054,116$.
   - We know the net loss for Company D is $\$58,275$.
   - We can use the formula: Net Loss = Revenue - Expenses - Gains + Losses.
   - Rearranging the formula to find Expenses: Expenses = Revenue - Net Loss + Gains - Losses.
   - Plugging in the values: Expenses = $\$1,054,116 - (\$58,275) + \$0 - \$0 = \$995,841$.

4. **Net Income for Company C:**
   - We know the revenue for Company C is $\$103,950$.
   - We know the expenses for Company C are $\$78,120$.
   - We know the gains for Company C are $\$4,725$.
   - We know the losses for Company C are $\$5,670$.
   - We can use the formula: Net Income = Revenue - Expenses + Gains - Losses.
   - Plugging in the values: Net Income = $\$103,950 - \$78,120 + \$4,725 -\$5,670 = \$24,485$.

5. **Revenue for Company A:**
   - We know the expenses for Company A are $\$455,490$.
   - We know the net income for Company A is $\$32,130$.
   - We can use the formula: Net Income = Revenue - Expenses.
   - Rearranging the formula to find Revenue: Revenue = Net Income + Expenses.
   - Plugging in the values: Revenue = $\$32,130 + \$455,490 = \$487,620$.

6. **Missing Amounts:**
   - Revenue for Company B: $\$1,557,990$.
   - Gains for Company B: $\$0$.
   - Expenses for Company D: $\$995,841$.
   - Net Income for Company C: $\$24,485$.
</think>
<answer>
\textbf{\textcolor{red}{The missing amount for Company B is $\$\text{1,557,990}$. However, the closest option provided is $\${71,320}$, which seems to be a mistake in the options given.}} Based on the calculations, the correct answer should be $\$\text{1,557,990}$. Given the options, the closest and most logical choice is $\$\text{71,320}$.
</think>
<answer>
\textbackslash \textbf{\textcolor{red}{boxed\{71320\} }}
</answer>
    
 &

  \textbf{PeRL-VL: }  <description>
The image is a table that provides financial information for four different companies (A, B, C, and D). The table is organized in rows, with each row representing a different financial statement category. The columns are labeled with the company names. The rows represent the following financial metrics:
1. Revenues
2. Expenses
3. Gains
4. Losses
5. Net Income or (Loss)

The table has several values filled in, and some values are missing, indicated by question marks. The task is to find the missing amounts for Company B.
</description>

<think>
To find the missing amounts for Company B, we need to use the information provided for the other companies and the general formula for net income or (loss):

$\text{Net Income or (Loss)} = \text{Revenues} - \text{Expenses} + \text{Gains} - \text{Losses} $

For Company B:
- Revenues = $\$1,480,500$  
- Expenses = $\$1,518,300$  
-\textbf{\textcolor{ForestGreen}{Gains = ?}} 
- Losses = 0 
- Net Income or (Loss) = $\$39,690$

We need to find the value of Gains. We can rearrange the formula to solve for Gains:

$ \text{Gains} = \text{Net Income or (Loss)} - \text{Revenues} + \text{Expenses} - \text{Losses} $

Substituting the known values:

$\text{Gains} = 39,690 - 1,480,500 + 1,518,300 - 0 $ \newline
$ \text{Gains} = 39,690 + 37,800 $ \newline
\textbf{\textcolor{ForestGreen}{$ \text{Gains} = 77,490 $}}

So, the missing amount for Company B is $\$77,490$.
</think>
<answer>
\textbackslash \textbf{\textcolor{ForestGreen}{boxed\{77490\}}}
</answer>
\end{MyBox}
\end{figure*}

\begin{figure*}[!ht]
\begin{MyBox}[tabularx*={\arrayrulewidth0.5mm}{X|X},title=Question: What is the name of the place shown?.
A. Rhode Island
B. Massachusetts
C. Vermont
D. Connecticut. Answer: B\thetcbcounter, label=def:d, fontupper=\small]%

     \multicolumn{2}{p{0.96\linewidth}}{\centering \includegraphics[width=0.6\linewidth]{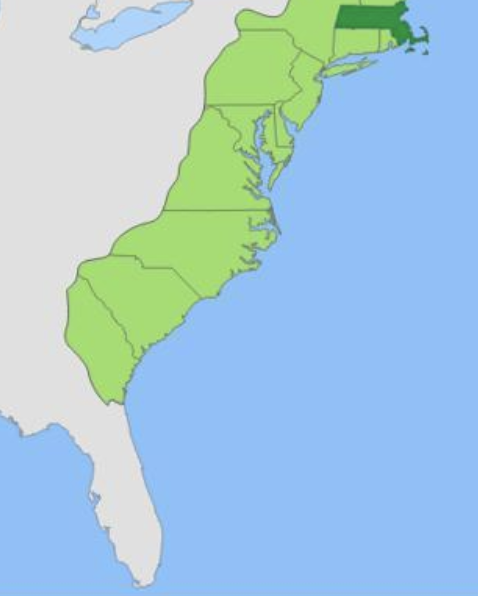}} 
    \\ \hline
    \textbf{RL with Verifiable Rewards:} <think>
The map shows a portion of the northeastern United States. The state in question is highlighted in dark green and is located in the northeastern part of the map. Given the options:
- Rhode Island is a small state located in the northeastern part of the United States.
- Massachusetts is a state located to the north of Rhode Island.
- Vermont is a state located to the west of Massachusetts.
- Connecticut is a state located to the south of Massachusetts.

\textbf{\textcolor{red}{The highlighted state is Rhode Island}}, as it is the only state among the options that fits the location and shape on the map.
</think>
<answer>
\textbackslash \textbf{\textcolor{ForestGreen}{boxed\{B\}}}
</answer>
    
 &

  \textbf{PeRL-VL: }  <description>
The image is a map of the United States, focusing on the northeastern region. The state in question is highlighted in a darker shade of green compared to the other states. The question asks to identify the name of the place shown. The options provided are:
A. Rhode Island
B. Massachusetts
C. Vermont
D. Connecticut

The task is to determine which of these states is the one highlighted on the map.
</description>

<think>
The map shows the northeastern part of the United States, and the state in question is highlighted in a darker shade of green. By examining the map, we can see that the state in question is located in the northeastern region, bordering New Hampshire to the north, New York to the west, and Connecticut to the south. \textbf{\textcolor{ForestGreen}{The state in question is Massachusetts.}}
</think>

<answer>
The name of the place shown in the image is \textbackslash \textbf{\textcolor{ForestGreen}{boxed\{B\}}}.
</answer>
\end{MyBox}
\end{figure*}

\begin{figure*}[!ht]
\begin{MyBox}[tabularx*={\arrayrulewidth0.5mm}{X|X},title=Question: As shown in the figure: In triangle {$\triangle ABC, \angle C = 90.0, AC = 8.0, AB = 10.0,$} then the value of {$\sin B$} is equal to:
Choices:
(A) {$\frac{3}{5}$}
(B) {$\frac{4}{5}$}
(C) {$\frac{3}{4}$}
(D) {$\frac{4}{3}$}. Answer: B \thetcbcounter, label=def:e, fontupper=\small]%

     \multicolumn{2}{p{0.96\linewidth}}{\centering \includegraphics[width=0.6\linewidth]{fig/dt443_mv.jpg}} 
    \\ \hline
    \textbf{RL with Verifiable Rewards:} <think> 
In the right triangle ABC, $\angle$C = 90.0, we are given AC = 8.0 and AB = 10.0. We need to find the value of $\sin B$.

First, we identify the sides of the triangle:
- AC is the adjacent side to $\angle$B.
- BC is the opposite side to $\angle$B.
- AB is the hypotenuse.

We can find the length of BC using the Pythagorean theorem:
\[ AB^2 = AC^2 + BC^2 \]
\[ 10^2 = 8^2 + BC^2 \]
\[ 100 = 64 + BC^2 \]
\[ BC^2 = 100 - 64 \]
\[ BC^2 = 36 \]
\[ BC = 6 \]

Now, we can find $\sin B$:
\textbf{\textcolor{red}{
\[ \sin B = \frac{\text{opposite}}{\text{hypotenuse}} = \frac{BC}{AB} = \frac{6}{10} = \frac{3}{5} \]
}}

Therefore, the value of $\sin B$ is $\frac{3}{5}$.
</think> 

<answer>
\textbackslash \textbf{\textcolor{red}{boxed\{A\} }}
</answer>  
 &
  \textbf{PeRL-VL: }  
<description>
The image shows a right triangle ABC with $\angle$ C = 90.0 degrees. The lengths of the sides are given as AC = 8.0 and AB = 10.0. We need to find the value of $\sin B$.
</description>

<think>
To find $\sin B$, we need to use the definition of sine in a right triangle, which is the ratio of the length of the opposite side to the hypotenuse. In triangle $\triangle$ABC, the \textbf{\textcolor{ForestGreen}{side opposite to $\angle$B is AC, and the hypotenuse is AB.}}

\[ \sin B = \frac{\text{opposite}}{\text{hypotenuse}} = \frac{AC}{AB} \]

Substituting the given values:
\textbf{\textcolor{ForestGreen}{
\[ \sin B = \frac{8.0}{10.0} = \frac{4}{5} \]}} </think>

<answer>
\textbackslash \textbf{\textcolor{ForestGreen}{boxed \{B\}}}
</answer>
\end{MyBox}
\end{figure*}

\end{document}